\documentclass[sigconf,nonacm,11pt]{acmart}
\AtBeginDocument{%
  \providecommand\BibTeX{{%
    \normalfont B\kern-0.5em{\scshape i\kern-0.25em b}\kern-0.8em\TeX}}}

\graphicspath{{fig/}{./}}

\copyrightyear{2021}
\acmYear{2021}
\setcopyright{rightsretained}

\acmConference{arxiv}
\acmDOI{Data and Visual Analytics}

\usepackage{listings}
\usepackage{subcaption}

\begin{document}

\title{RtFPS: An Interactive Map that Visualizes and Predicts Wildfires in the US}


\author{Yang Li,
    Hermawan Mulyono*,
    Ying Chen*,
    Zhiyin Lu,
    Desmond Chan
}
\thanks{* Equal contribution}
\affiliation{%
\institution{\textbf{Georgia Institute of Technology}}
}

\affiliation{\{yli3297, hmulyono3, ychen3267, zlu335, cchan71\}@gatech.edu}

\begin{abstract}
Climate change has largely impacted our daily lives. As one of its consequences, we are experiencing more wildfires. In the year 2020, wildfires burned a record number of 8,888,297 acres in the US\cite{nifc07}. To awaken people's attention to climate change, and to visualize the current risk of wildfires, We developed RtFPS, "Real-Time Fire Prediction System". It provides a real-time prediction visualization of wildfire risk at specific locations base on a Machine Learning model. It also provides interactive map features that show the historical wildfire events with environmental info. Source code:\\ \color{magenta}{\url{https://github.com/yangland/rtfps}}

\end{abstract}


\maketitle

\section{Introduction}
{\bfseries} 
Climate change is arguably the most imperative crisis of our century. Studies have been conducted throughout decades to identify its impact on the environment and potential connections with extreme natural phenomena, including the rising frequency of wildfires observed in many parts of the world\cite{McCaffreySarah2015C2-S,WesterlingA.L2006WaES,FriedJeremyS2008Pteo,Westerling2008,J.R.Marlon2009WRtA,DoerrStefanH2016Gtiw}. While research results have been abundant, the public is still largely uninformed of the imminent danger due to poor communication. Therefore, our group aims to visually convey the significance of the issue by creating an interactive map that show a real-time forecast of the wildfire and demonstrates how climate change has been affecting the occurrence of wildfire in the United States from 1992 to 2015\cite{Karen2017,XieWeiwei2020DCAb,DraxlCaroline2015TWIN,DennisonPhilipE2014Lwti}. We provide predictions of wildfire risks base of our Machine Learning model using weather and fuel conditions. We also present the change of climate with data in temperature, precipitation, vegetation, etc. These climate data are then overlapped with a visualization of wildfire events on the map to exhibit their correlations\cite{LabosierChristopherF2015Wtco,WesterlingAL2011Ccag,CoenJ2015WF|W,KalabokidisKostas2016Aawp,2011Favo,BallantyneAnneGammelgaard2016Iocc,J.R.Marlon2009WRtA}. 



We introduced several new features in this project. The \textbf{time slider} is expected to help the audience see the evolution of wildfires over time more clearly. We aim to produce a \textbf{simple fire risk model} which depends on data that is abundantly available: weather, wind, and fuel.
We also introduced a methodology to \textbf{work with imperfect data}, filling the missing and join the data creatively to build the fundamental dataset from visualization and training. 
We found the severity prediction with a \textbf{machine learning technique}, which outputs a real-time forecast of the wildfire risk. These features achieved our goal to educate our audience and provide more insights into wildfires in the US.

Our visualization project is the first among many wildfire visualization projects that presents wildfire events with many climate attributes on the same map. We are also the first to implement an interactive time bar function for users to select the time point that they are interested in. Our work is also the first to attempt machine learning prediction as one of the components of visualization. We believe that these three innovations help educate our audience regarding wildfire and climate change, and provide them with an insight into the relationship between the occurrence of wildfire and weather conditions.

\section{Related Work}
{\bfseries}
There is no simple yet comprehensive wildfire visualization available today. Although many projects have been completed, many of them are either unable to explain the relationship between climate change and wildfire \cite{CRAWL20172230} or too complicated to be easily understood by the
public \cite{AMATULLI20060817}. There is also limited work to explore efficiency and understandable way to make a prediction for wildfire which can provide an easy tool for the public. 

We also researched the current machine learning methods to predict wildfire occurrence.
Most of the other studies listed the machine learning models and explained each of them by analyzing the result of performance metrics \cite{MaJun2020Rdow} \cite{SayadYounesOulad2019Pmow}.  They tend to dive deep to explain the prerequisite and algorithms of models, validating these models using cross-validation methods, focusing more on the comparing models\cite{AshimaMalik2021DWRP}. None of them try to implement real-time predictions which is the core application for this research.

Our visualization on historical data shows not only wildfire events but also changes in climate in the region on the same map. This would enable the audience to infer from the map the relationship between climate change and the occurrence of wildfire in the given time frame. Moreover, the map will be interactive by including a time bar for the users to drag on. The users could choose to pause at one specific time point or drag along the bar to read the map dynamically. This gives the users an easy and effective way to explore the issue on their own and figure out the message that we are trying to convey from the map.




{\bfseries}

{\bfseries} 
 Our primary audience is people who are yet to be educated about the potential impact of climate change. The more specific users of the interactive map would include researchers studying the issue, policymakers in regions with forested areas, local habitats in the areas where wildfire is a potential threat, government analysts, firefighters, etc. These groups of people will benefit from more insights into wildfires\cite{ScottThompsonCalkin2013}.

\section{method}

\subsection{Data Sources}
Wildfire occurrence and behavior are affected by many factors including temperature, wind, vegetation, fuel availability, fire regime, landscape, etc\cite{LabosierChristopherF2015Wtco,WesterlingAL2011Ccag,CoenJ2015WF|W}. The data we used to build the wildfire visualization map consists of three major categories: weather(temperature, precipitation, and wind), fuel, and fire.


\subsubsection{Temperature and Precipitation}
Data for weather contains rain and temperature information collected from the GHCN (Global Historical Climatology Network)-Daily, which is an integrated database of daily climate summaries from land surface stations across the globe. GHCN-Daily contains records from over 100,000 stations in 180 countries and territories. It provides daily variables, including maximum and minimum temperature, total daily precipitation, snowfall, and snow depth. We chose the precipitation and temperature from all available variables provided and also preserved the location information for the mapping later. 

\subsubsection{Wind}
The National Oceanic and Atmospheric Administration (NOAA) spanned our desired time frame of 1992-2015
with daily updates, it contained the wind vector data indicated by its u and v components. To calculate wind speeds, we researched mathematical instructions and used the following formula (1) to convert u and v components of the winds into absolute wind speeds. 
 \begin{equation}
    wind speed=\sqrt{u^2+v^2}
 \end{equation}

The original .nc files obtained from the NOAA website were translated into .csv files using JupyterLab. The processed data contains time, latitude, and longitude information, and calculated wind speed. The overall quality of the data is great with no need to deal with missing data points.

\subsubsection{Fuel}
Natural combustibles dead and live vegetation affect the severity of wildfire. In particular, fuel flammability is indicated by Fuel Moisture Content (FMC), which is detailed explained in Yebra's paper\cite{Yebra2018}. Previous efforts to extract fuel moisture data from the National Fuel Moisture Database (NFMD) have been completed by a research group from Stanford University. Their published codes could be found at \href{https://github.com/kkraoj/lfmc_from_sar}{here}~\cite{Rao2020}. Motivated by a similar demand for such data, we consulted their codes and modified the codes based on the nature of our project. We used a Google Chrome driver to automatically submit download requests for all available data. The obtained data were then consolidated into a single \lstinline{csv} file containing information of the time, fuel type, fuel flammability, and geographic coordinates. 

\subsubsection{Wildfire}

Fire incidents data was extracted from a dataset from Kaggle in SQLite format and was performed with basic error-checking. Redundant records were also identified and removed. The pruned data were then read as a pandas data frame and useful columns were selected. The latitude and longitude information
can be used to associate data points with locations on a map. 

\subsubsection{Real-time Weather Data}
The real-time weather data are from weatherstack API. The interface provides both historical and forecast data to be extracted. We get temperature, precipitation, and wind data from API. By combining 14 days of historical data and 7 days of forecast data, We set up a dataset of a total of 21 days' weather records for feeding our machine learning model, through which we make predictions for wildfires.

\subsection{Data Processing}
After collecting and processing the data using the methods described above, we combined the data for each feature into one dataset by translating location information into FIPS location units. FIPS stands for Federal Information Processing Standards, and the 5-digit codes uniquely identify geographic areas in the United States. The first 2 digits tell the state, and the last 3 digits indicate the county. The integrated dataset was then used as the training and testing sets of a wildfire prediction model.

There are two types of fuel moisture data: dead and live. One fuel sample site could have a data type of one or both. The fuel moisture content(FMC) for the dead is always much less than the live fuel, about 15\% to 30\% of the live fuel. The model should not merge live and dead FMC as one feature column.

Ideally, we would like each location unit (FIPS) to have monitoring sites for all temperature \& precipitation(TP), wind, and fuel, and provide us data every date from 1992 to 2015. However, this is not the case in both time and space measurements. Firstly, fuel data have a very low sampling frequency. Usually,  the interval of fuel record is between a few weeks or even months for one location.
As for TP, some of the TP sites have a very short data period, there are 967 TP sites(total 2041) that cover less than 50\% of the time coverage between 1992 and 2015.

One FIPS location unit usually does not have all three monitoring stations in its domain. Most FIPS units only have one of the three station types. The differences in the site's coverage and data density make the environmental data hard to join together with the FIPS frame.


Below are the sites numbers for the three stations types in US.

\begin{itemize}
    \item Temperature \& precipitation: 2041
    \item Wind: 311
    \item Fuel: 1250
    \item FIPS: 3218
\end{itemize}

In NIFC wildfire data, about 2/3 of the wildfire records are missing at least one of the state or the county FIPS codes. The wildfire data include the columns for the fire's start date and end date, but 47.1\% records are missing the end date.

To make up the missing data and make data suitable to join based on the location FIPS code and date, we need to process data based on reasonable logic:

\begin{enumerate}
     \item Assign TP, wind, and fuel sites to each FIPS unit based on the distance calculated from the center of the FIPS unit and the site location. Select the closest site for each data type as assigned sites for each FIPS unit, as shown in Figure~\ref{fig:assign_sites}.
        \begin{figure}[h]
        \centering
        \includegraphics[width=7cm]{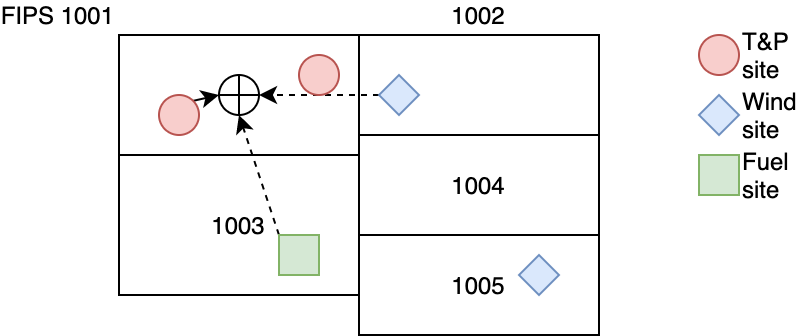}
        \caption{Assign sites to FIPS based on distances}
        \subcaption*{Each FIPS unit map to the site within its territory or the closest one based on distance}
        \label{fig:assign_sites}
        \end{figure}
    \item Incorporate the central longitude and latitude for each FIPS unit.
    \item Use only the live fuel moisture data as FMC. The national fuel moisture database does not provide the weight between living and dead for each site. The dead fuel data makes up a much smaller percentage (21.7\%) of all 262679 data records for fuel. By using only the live fuel data, our fuel indicator is consistent.
    \item Adding the missing data for fuel. The data between two dates would be filled in with the FMC value from the earlier date. The period between the first available date and 1992-01-01 is filled with the FMC of the first available date. The period after the last available date and 2015-12-31 is filled with FMC of the last available date. The whole adding process shown in Figure~\ref{fig:add_missing_fuel}. After the filling, all fuel stations would have FMC value from 1992 to 2015.
        \begin{figure}[h]
        \centering
        \includegraphics[width=7cm]{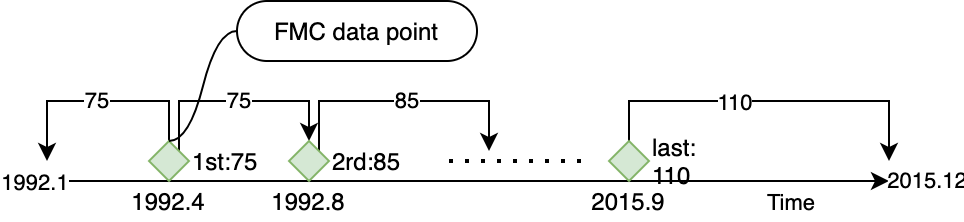}
        \caption{Add missing data for fuel}
        \subcaption*{Date between 2 FMC records uses the value from the previous date}
        \label{fig:add_missing_fuel}
        \end{figure}
    \item For TP sites, we use only the 929 sites that have larger than 50\% coverage on dates between 1992 - 2015. After this setup, the average distance from FIPS to its TP site increases from 95km to 171km, but it provided better data coverage on available dates. Comparing with the average wind site distance of 126.21 km, the average distance of TP to the center of its FIPS unit is still reasonable.
    \item Cleaning up data by removed the extreme records of temperature records which are beyond the value of world records. We only use temperatures between -90 and 100 °C.
    \item Assign FIPS code for all wildfire records. A FIPS searching program is created to finds the FIPS base on the incident reported longitude and latitude, and also validate the result with the original state and county FIPS in cases if we have them from the record.
    \item For the wildfire events that do not have an end date, we use the start date as the end date. The fires without an end date have an average area burned much smaller (24.88 acres compare with the average 74.52 acres). The majority of all fire incidents(90.28\%) only last for one day. It is a reasonable assumption that these fires lasted for less than 24 hours.
    \item Convert the event records to an aggregated fire size burned on the date and fips. The original fire data are incident event records. For one date, at one FIPS location, there are could be more than one fire incident. We averaged the fire size to each date between the fire start date and the end date. Then aggregate the sum of the fire sizes of all fire incidents on that date, group by FIPS and date (Figure~\ref{fig:convert_fire}). Once converted, the fire size daily sum shows the size of fire burned each date on all FIPS units.
        \begin{figure}[h]
        \centering
        \includegraphics[width=7cm]{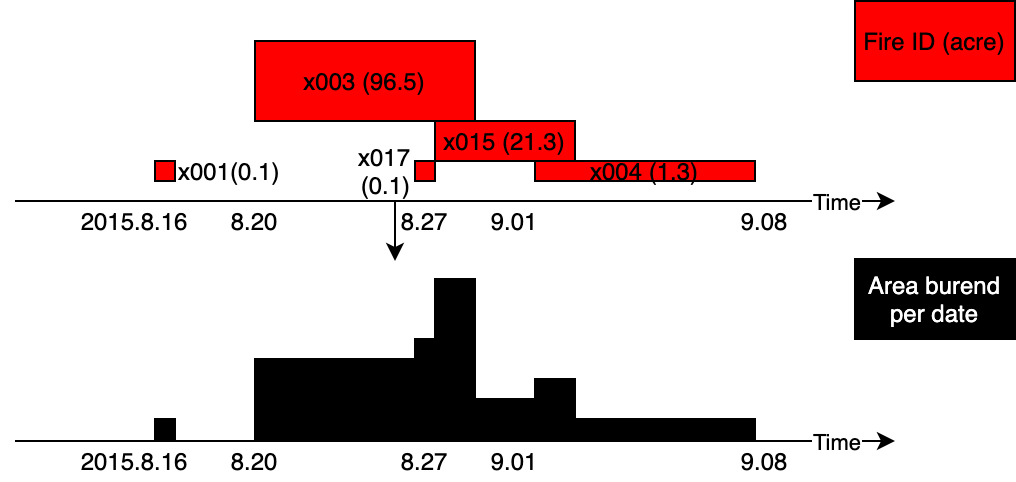}
        \caption{Convert fire event records to aggregated daily fire burned area}
        \subcaption*{Fire sizes from different events are projected to the Time axis for the per date area sum}
        \label{fig:convert_fire}
        \end{figure}
\end{enumerate}
After the data processing, we are able to join the environmental data with the wildfire data on columns of FIPS and date. The joined dataset contains the following columns:
\begin{itemize}
    \item fips
    \item longitude
    \item latitude
    \item date
    \item wind
    \item tmax
    \item tmin
    \item tavg
    \item fmc
    \item prcp
    \item fire size day sum
\end{itemize}



The last data process step produces a dataset for fitting in our predicting model. We need the same real-time X features to generated prediction on the trained model. The weather API needs location information such as coordinates, cities as the query input. It is not feasible to call API for all coordinates or cities because of the large amount of data and API limitations. We selected a list of major US cities as prediction locations. We feed these locations to Google Geocoding API, find their cities and counties information for incorporating the fuel data. Weather information and fuel FMC are then combined joined by cities for the prediction dataset.

\subsection{Machine Learning Model}

We discover the correlation between features and the fire size. We then further test different regressor models and select the best performer.

\subsubsection{Feature Engineering}
In the first attempt of fitting the data into a machine learning model, we use the "fire size day sum" as y, and the rest features besides the "date" as X. The results from machine learning models are not ideal, random forest regressor and neural network regressor shows negative R2 scores.

\begin{figure}[h]
        \centering
        \includegraphics[width=7cm]{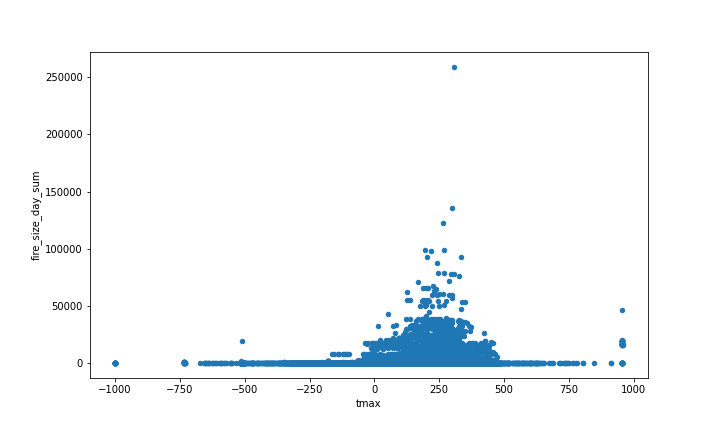}
        \caption{Maximum temperature and fire daily sum (Unit 0.1°C}
        \subcaption*{10 - 35°C range's fire occurs most frequently}
        \label{fig:t_f}
        \end{figure}

\begin{figure}[h]
        \centering
        \includegraphics[width=7cm]{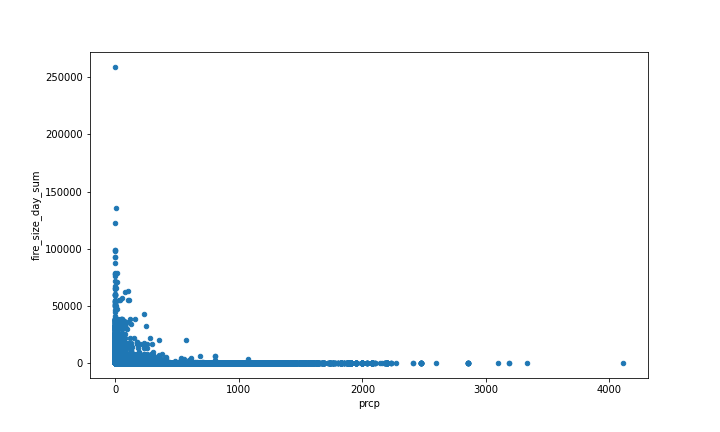}
        \caption{Precipitation and fire daily sum}
        \subcaption*{Dry weather with little or no rain has a much higher fire records}
        \label{fig:p_f}
        \end{figure}

\begin{figure}[h]
        \centering
        \includegraphics[width=7cm]{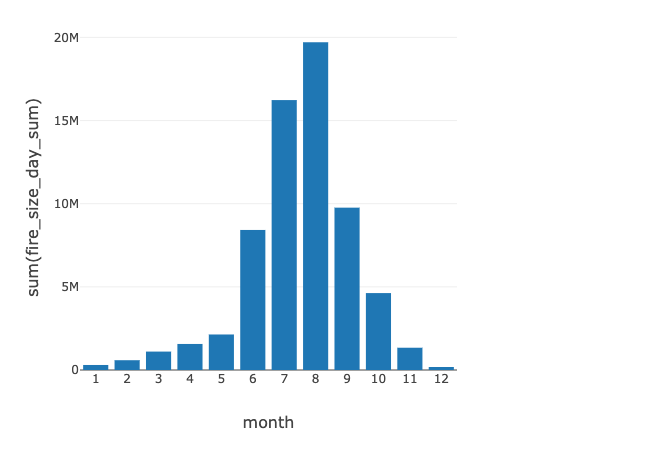}
        \caption{Aggregated sum by month}
        \subcaption*{The summer months June to Sep have the majority of the fire records}
        \label{fig:m_f}
        \end{figure}

However, we can see from the Figure~\ref{fig:t_f}, ~\ref{fig:p_f}, ~\ref{fig:m_f} that the wildfires do have correlation relationship with tmax, prcp and month.

We believe that the low accuracy of the daily data feature model is due to randomness. We observed that one location could have a very small or zero fire burned area even when the environmental condition is in favor of fire on that date. Wildfires were caused by different reasons, but the incident to start a fire may not happen on that specific date in that specific location. So the "safety" shows from a low fire burned area value is not actually mean the risk is low on that date in that location.



\begin{figure}[h]
        \centering
        \includegraphics[width=7cm]{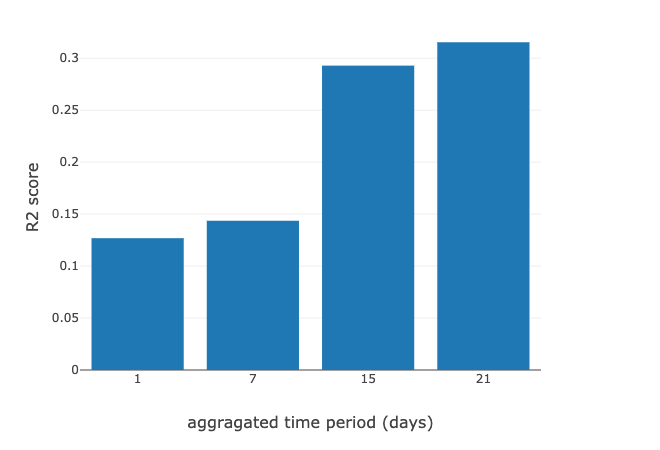}
        \caption{Gradient Boosting Regressor R2 scores with different aggregated time periods}
        \subcaption*{Extending the aggregated time period significantly improves the R2 scores}
        \label{fig:agg}
        \end{figure}

To remove the random factor and show up the statistical significance, we engineered the features to be aggregated in a 21 days window. The X features are averaged environmental variables, and they became the sum of fire burned area in the 21 days. The model's accuracy improved significantly. We use R2, coefficient of determination, to measure the accuracy of the regressor. The R2 score responding to an aggregated period from 1 to 21 days shown in Figure ~\ref{fig:agg}.

\subsubsection{Learning Algorithms}
The y value that we are predicting is the fire size day sum for a location. It is a continuous value, base on this we tried multiple machine learning regressor. 
\begin{itemize}
    \item Linear Regressor
    \item Random Forest Regressor
    \item K Neighbors Regressor
    \item Neural Network Regressor
    \item Gradient Boosting Regressor
\end{itemize}

\begin{figure}[h]
    \centering
    \includegraphics[width=8cm]{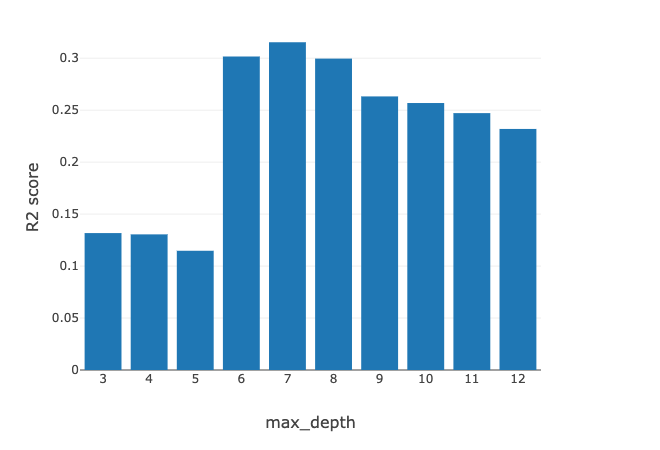}
    \caption{Gradient Boosting Regressor hyper-parameter tuning (max depth)}
    \subcaption*{Max depth = 7, best R2 score = 0.315}
    \label{fig:gbr}
    \end{figure}

After turning hyper-parameters for all machine learning models, Gradient Boosting Regressor achieved the best r2 score of 0.315, with n estimators=150, learning rate=0.01, and max depth = 7. Figure ~\ref{fig:gbr} shown R2 scores corresponding with different max depth parameters. Figure ~\ref{fig: regressor} shown the best R2 scores from all four machine learning regressors.

    \begin{figure}[h]
    \centering
    \includegraphics[width=8cm]{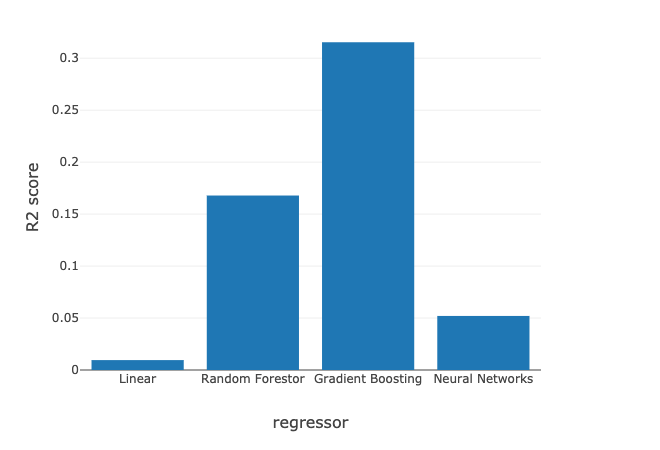}
    \caption{Comparison of 4 learning methods of regression}
    \subcaption*{Gradient Boosting out performs the other 3 regressors}
    \label{fig:regressor}
    \end{figure}

The Linear regressor is a good indicator if feature engineering was effective. When using the daily features, the score of the linear regressor was around 0.001. After adding features of month, long and latitude, and increase the sampling window from 1 day to 21 days, the score of linear regressor also increased to 0.00953. We are not using linear regressor as our selected model since the accuracy was still too low. But as the linear regressor score improves we can also achieve a higher accuracy score from the other more complex regressor models.

\subsubsection{Machine Learning result presentation}

To provide an easy way for users to understand the meaning of the prediction of the machine learning model, we map the result to the US size of fire standards that indicates fire severity. Table~\ref{tab:plan} shows how it works. 

\begin{table}[h]
  \caption{Fire size classes}
  \label{tab:plan}
  \begin{tabular}{ccl}
    \toprule
    Fire Size Class&Fire size \\
    \midrule
    \colorbox{green}{\textbf{A}}&0-0.25 acres\\
    \colorbox{cyan}{\textbf{B}}&0.26-9.9 acres\\
    \colorbox{yellow}{\textbf{C}}&10.0-99.9 acres\\
    \colorbox{pink}{\textbf{D}}&100-299 acres\\
    \colorbox{orange}{\textbf{E}}&300 to 999 acres\\
    \colorbox{red}{\textbf{F}}&1000 to 4999 acres\\
    \colorbox{purple}{\textbf{G}}&5000+ acres\\
    \bottomrule
  \end{tabular}
\end{table}


There is a difference as the US fire standard is based on events of the fire incident, while the model output the sum of all fire burned areas. But we believe as the size of one FIPS unit is small enough, a standard like this would be helpful and intuitive for most readers.


\subsubsection{Conclusion of Analysis and Learning}
\begin{itemize}
    \item Wildfires happens much more often than people usually think. On average, for one FIPS unit, 2.1 fires are happening every month. 
    \item If the condition is in favor of fire, it is unlikely the place could hold as fire-free in a period of a few weeks. A large fire is unavoidable under bad environmental conditions. We have seen wildfires caused by vehicles and careless human activities as reported in news. But based on our study, these causes are not the main factors to blame for. The fire could happen in one way or the other.
    \item Wildfires prediction only has statistically significant in a macro sense. We are not able to predict how much area would be burned tomorrow at one county location. But based on the recent 3 week's weather conditions, we can predict and provide a useful indicator to show the risk of wildfire. 
\end{itemize}

\subsection{Data Visualization}




For the visualization, we will use D3 powered with Flask to show wildfire, climate, geographical data, and prediction results on our map. D3 is a popular JavaScript package for interactive visualization, which came natural to us to choose D3 for our interactive map. Flask is a very popular and simple web development Python library. We used D3 and Flask to set up a time bar that users can play with interactively. These tools were also used to show features on our map. For example, when the viewer drag the time bar to a specific time point, the counties would change to a specific color corresponding to its value, according to the features that the user selected. When the mouse hover over a county, a tip is shown to tell the county name.

\subsubsection{Model-View-Controller pattern}
\begin{figure}[h]
    \centering
    \includegraphics[width=7cm]{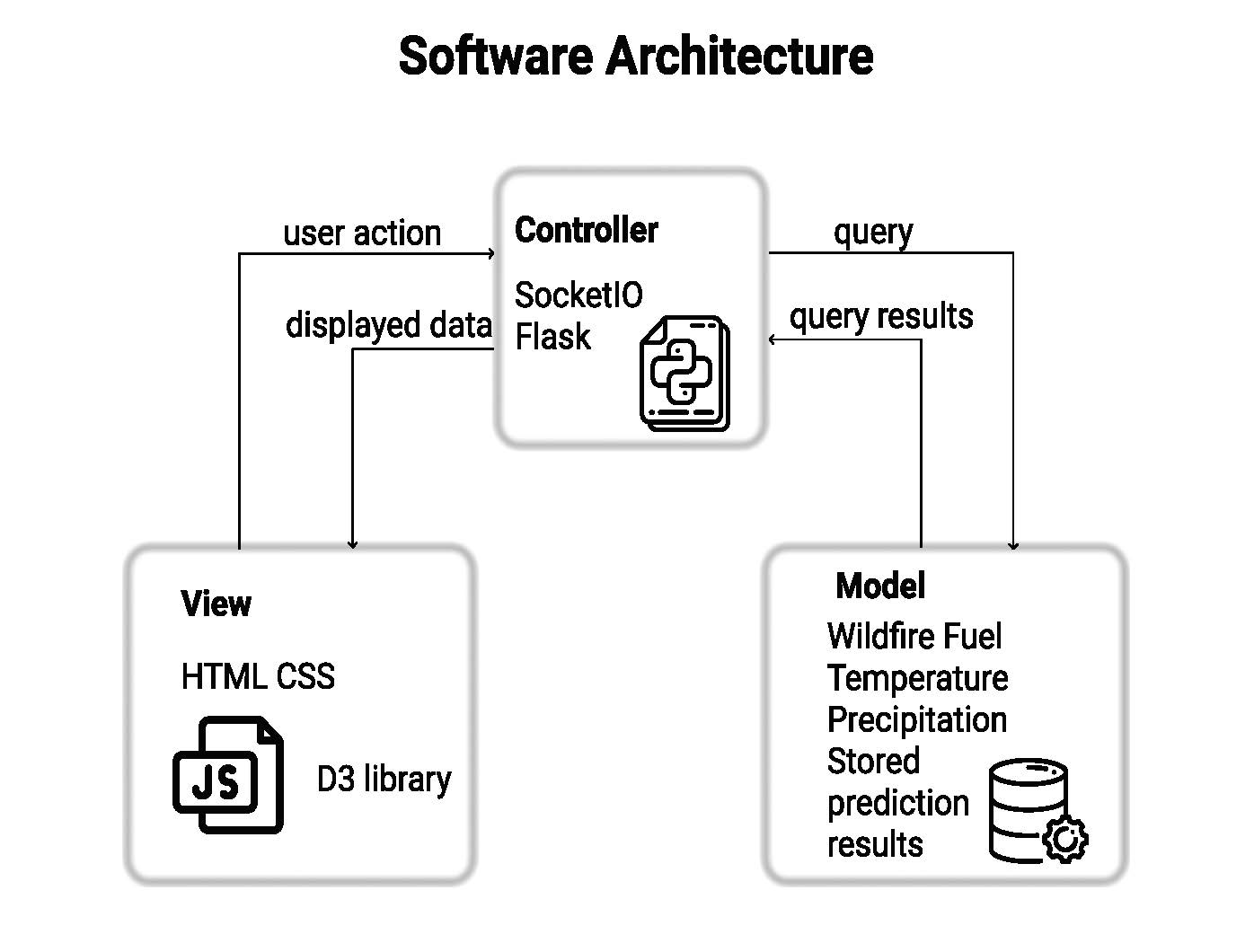}
    \caption{software architecture}
    \subcaption*{The data flow between View, Controller and Model}
    \label{fig:software architecture}
\end{figure}

Our web application structure used the Model-View-Controller pattern (or MVC). As is shown in Figure~\ref{fig:software architecture}, a user requests to view the wildfire map with desired features at a specific time point by sliding the time bar and tick relevant check box and radio buttons. The user action is then sent to the controller. The controller then invokes the model, where the database is stored and receives the query request. The query results are then passed back to the controller and finally updated in the view. To make it possible, we opted to use socket communication instead of HTTP GET so that users do not need to refresh their browser every time they perform an action. The database could be in any format that the model can read. In our case, We used CSV and SQLite for our convenience. 

\subsubsection{Web Visualization}

\begin{figure}[h]
    \centering
    \includegraphics[width=8cm]{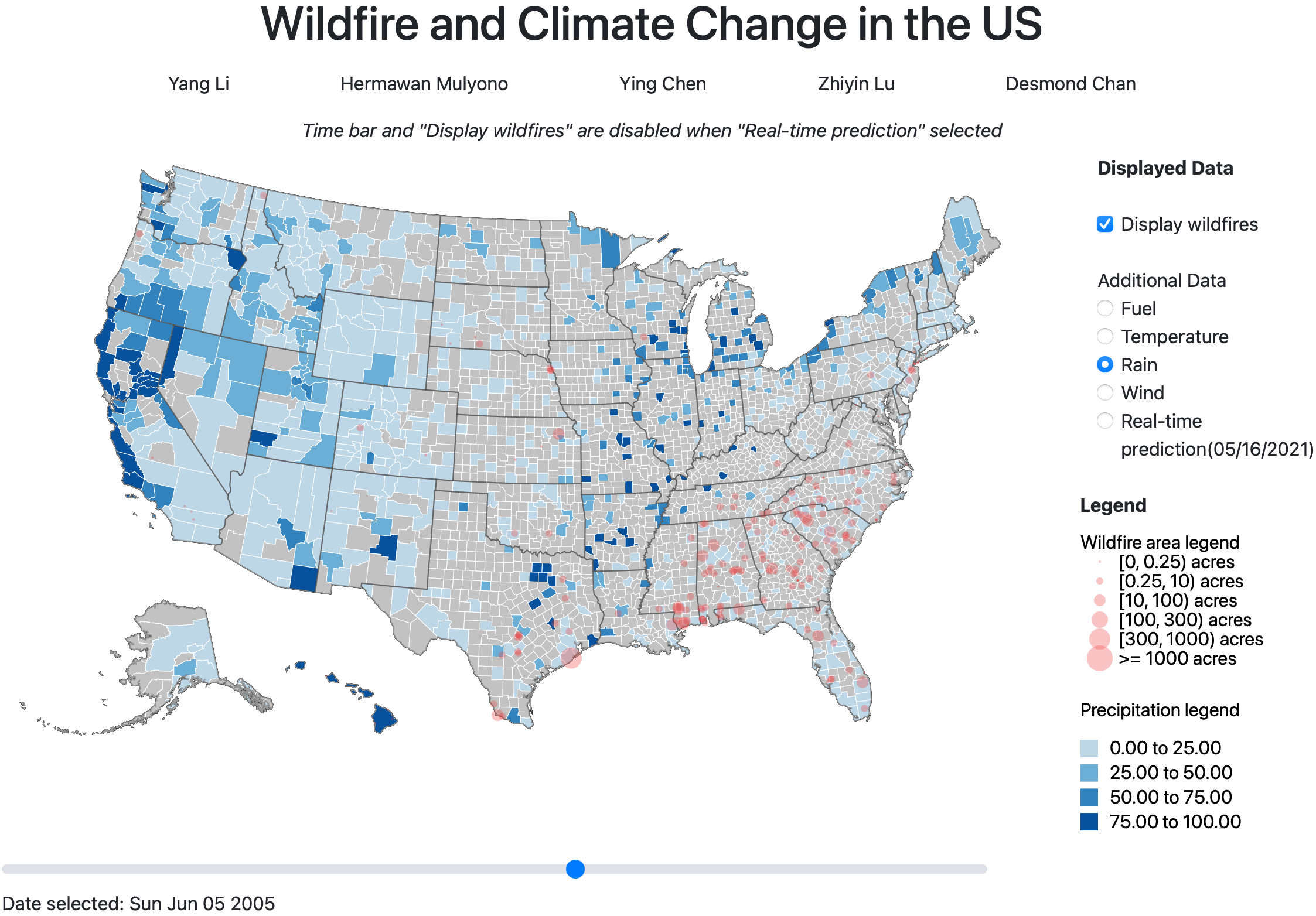}
    \caption{Visualization (wildfire with rain)}
    \label{fig:map}
\end{figure}

The final version of our visualization is shown in Figure~\ref{fig:map}. Users could pick any arbitrary date between 1992 and 2015 by scrolling the time bar at the bottom. They could also choose the features that they wish to see on the map by switching the options on the right-hand side of the map. The map receives the date selected, sends the query request, and then broadcasts the query result on the corresponding regions on the map. The climate features and prediction results are shown with colors of different scales on the choropleth map, and vegetation and wildfire data were presented as opaque circles with different radii. 


Wildfire events are displayed by default and could be removed by unchecking the "Display Wildfires" box above all other feature options. A red circle on the map indicates the occurrence of wildfire in that specific location on the map. The size of the circle tells the severity of the wildfire, defined by how many acres of forest and vegetation was burnt by that fire. Users could also click on the "machine learning prediction" option on the right, to see how predictions match the actual historical data.

An example demo scenario could be like this: a user wants to know the temperature on the date when a wildfire happened. The user could then scroll the time bar to find the time point and then click on the temperature option to check the data. The user may also like to see a prediction for the county, then the "machine learning prediction" option could be selected to see the result.

\subsubsection{Prediction Visualization}

When the user selects the "real-time prediction" radio button, the web app grabs the back-end file generated by the machine learning procedure. The machine learning procedure is scheduled to run once per day. It calls the Weatherstack API to fetch the previous 14 days' weather history and future 7 days' weather forecast. Averaging the X features for these 21 days and incorporate the latest fuel FMC, the program then feeds these data to a pre-trained machine learning model(Gradient Boosting Regressor) to get prediction results per FIPS unit on the map. 

Since the prediction is based on the current weather and fuel condition, "Time bar" and "Display wildfires" features are disabled when the "real-time prediction" radio button is selected.

Functions of the web application are shown in the appendix Figure~\ref{fig:vis_pred}, \ref{fig:vis_temp}, \ref{fig:vis_fuel}, \ref{fig:vis_wind}.

\section{Conclusions and Future work}

This paper introduced RtFPS, a novel system for visualizing wildfire risk and wildfire history in the US. RtFPS helps the user understand the current risk of wildfire and presents the historical wildfire events with temperature, precipitation, vegetation, and wind information. RtFPS achieves fire prediction by gathering real-time weather data through a third-party API and applying a machine learning model which learned from the historical wildfire events. Letter grades generated by the prediction model incorporate into the map provide the user an informational and easy-to-use interface to understand the danger of wildfire and climate change.


Looking forward, there are many ways to improve this project.

\begin{enumerate}
  \item The visualization and machine learning model could be designed in a way that they could continuously accept new data and generate new learning models. We did not get the data from 2015 to 2020. Which would be very helpful to improve the model. Ideally, the system could set up automated procedures to download the new data to update the visualizations and machine learning model. So that the model could be self-adjusted and updated monthly. The learning process could be adjusted that the model weights more on the more recent data.

  \item Wildfire is a global issue, we only had the data in the US and the project scope only covers the US. If possible the project could be extended to the whole world. So that we could learn from more data and cover other places such as Australia.
  \item The machine learning model accuracy might be improved after including more related data features such as air humidity, population, and more detailed FMC data of fuel.
  
  \item We did not deploy our visualization work onto a cloud server. If possible, we would explore more possibilities on deploying our work and experiment with many ways like Google Cloud to store our data remotely, or even structures other than MVC. So more people could try out RtFPS.
\end{enumerate}


\section{Authorship} 



YL conceived of the study. HM, YL, ZL, DC, retrieved data. HM designed the application structure and coded the web application. YL and ZL performed study in data analysis and machine learning models. YL implemented the data processing procedure and searched for the optimum Machine Learning models. YC coded the API data fetching and the prediction features. YL, HM, ZL, YC contributed to interpretation of the data and ideas of visualization.


\bibliographystyle{ACM-Reference-Format}
\bibliography{report-references}


\newpage
\onecolumn
\appendix
\section{Appendix}
\begin{figure}[ht]
    \centering
    \includegraphics[width=0.75\textwidth]{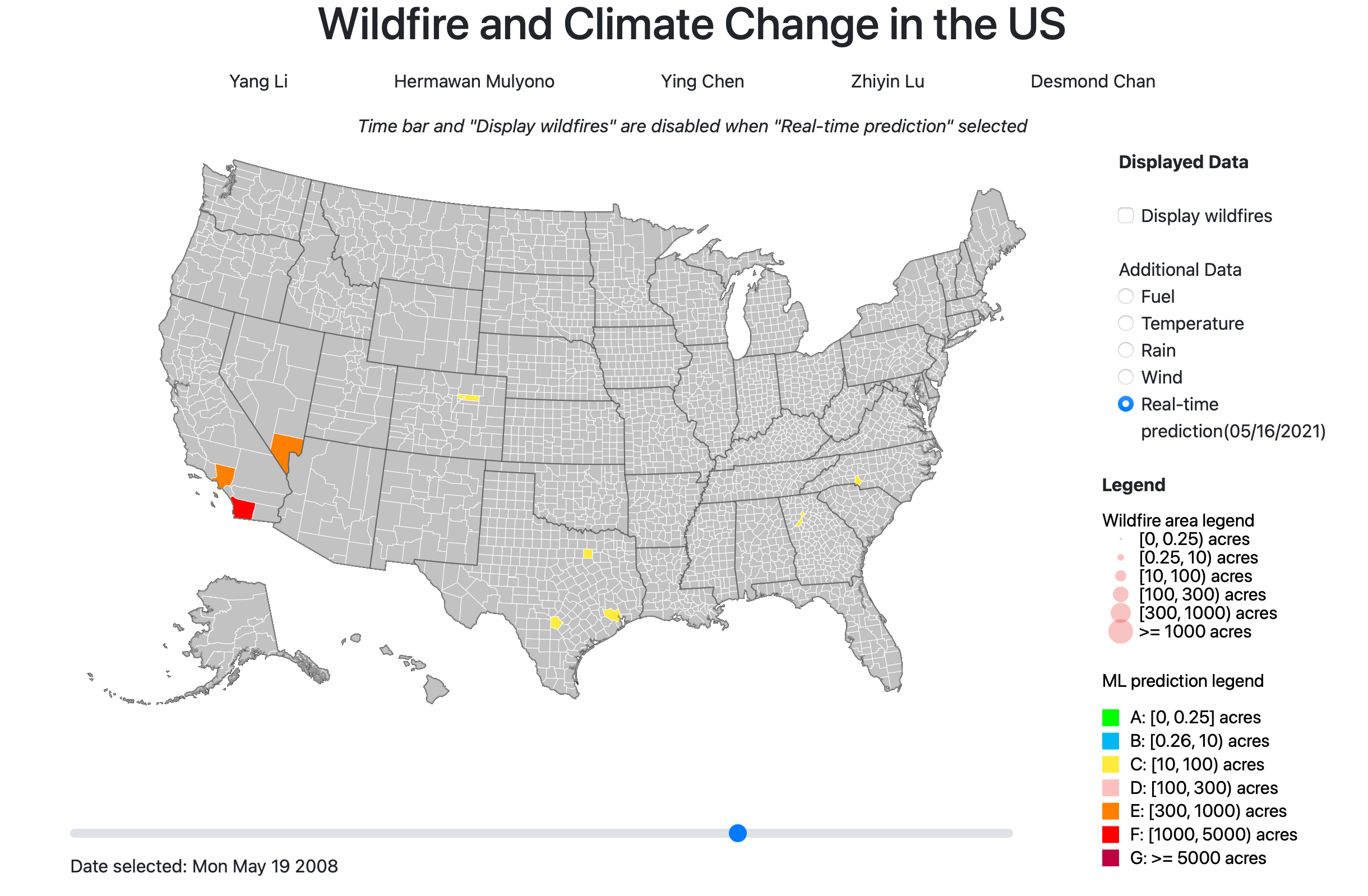}
    \caption{Visualization (Real-time Prediction)}
    \label{fig:vis_pred}
\end{figure}

\bigbreak

\begin{figure}[ht]
    \centering
    \includegraphics[width=0.75\textwidth]{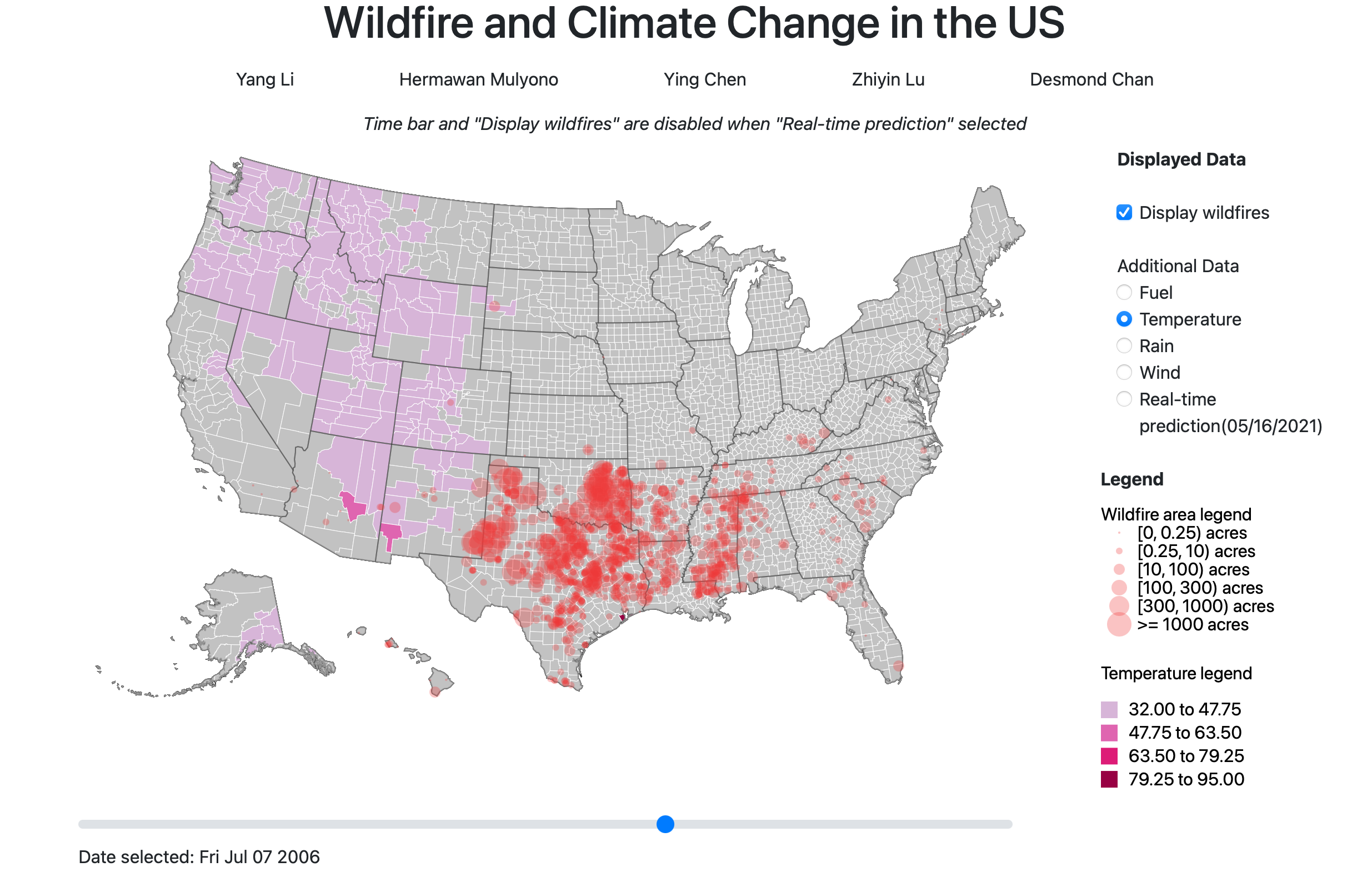}
    \caption{Visualization (wildfire and temperature)}
    \label{fig:vis_temp}
\end{figure}

\newpage

\begin{figure}[ht]
    \centering
    \includegraphics[width=0.75\textwidth]{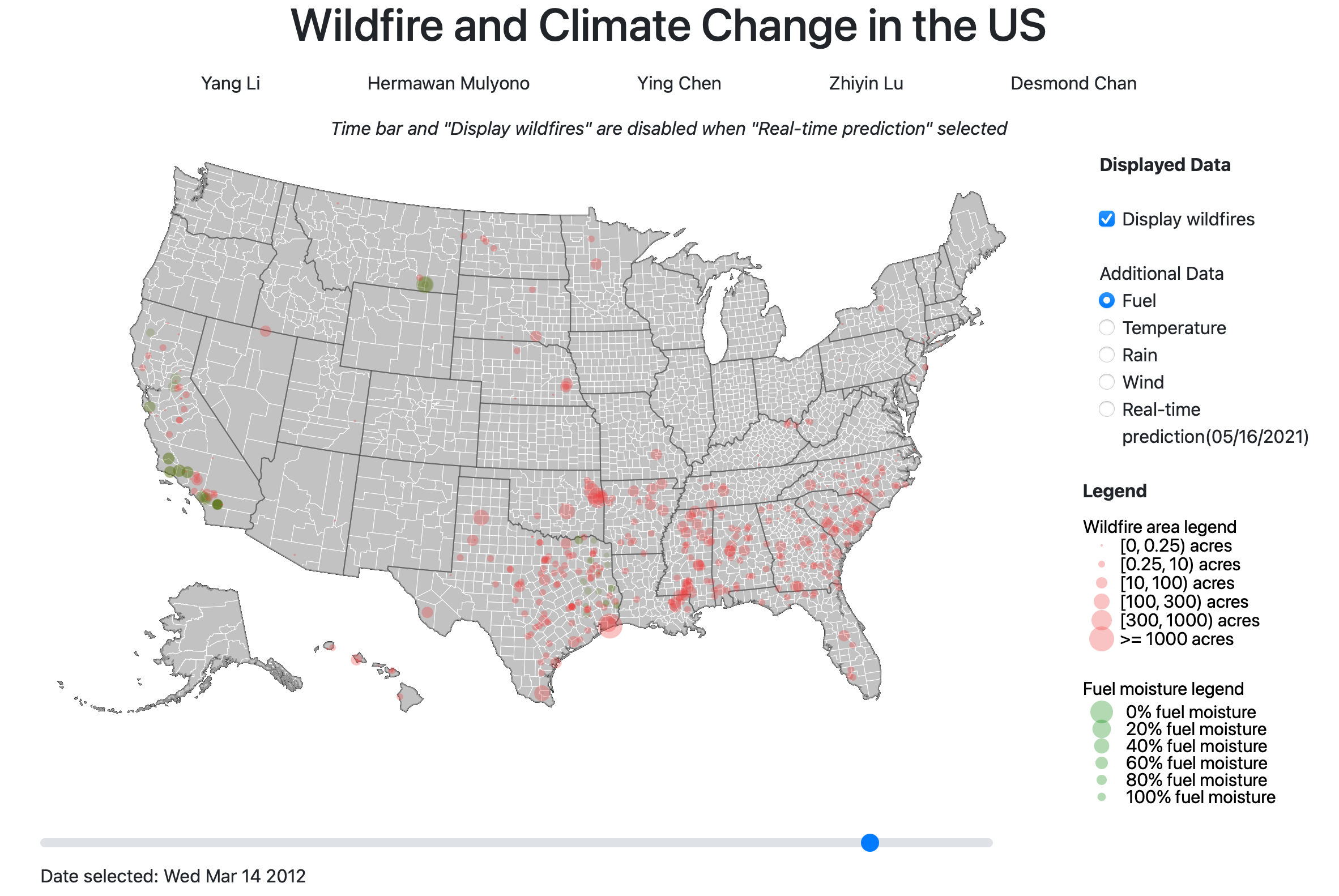}
    \caption{Visualization (wildfire and fuel)}
    \label{fig:vis_fuel}
\end{figure}

\bigbreak

\begin{figure}[ht]
    \centering
    \includegraphics[width=0.75\textwidth]{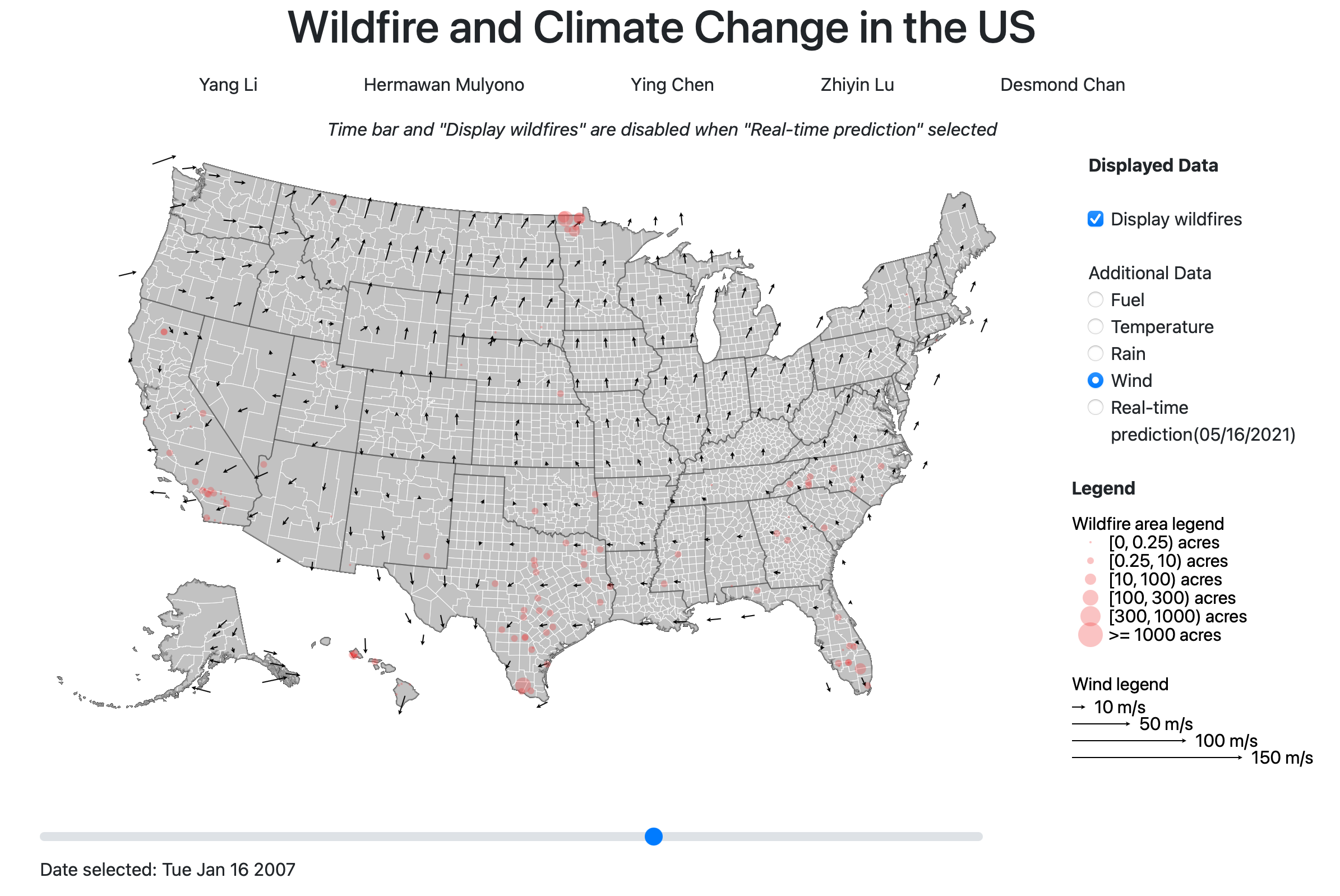}
    \caption{Visualization (wildfire and wind)}
    \label{fig:vis_wind}
\end{figure}

\end{document}